\documentclass[10pt,twocolumn,letterpaper]{article}

\usepackage{iccv}
\usepackage{times}
\usepackage{epsfig}
\usepackage{graphicx}
\usepackage{amsmath}
\usepackage{amssymb}
\usepackage{graphicx}
\usepackage{amsmath}
\usepackage{amssymb}
\usepackage{booktabs}

\usepackage[flushleft]{threeparttable}

\usepackage{colortbl}  
\usepackage{xcolor}
\usepackage{array}
\usepackage{pifont}
\newcommand*{\affmark}[1][*]{\textsuperscript{#1}}
\usepackage{bbding}
\usepackage{algorithm}
\usepackage[noend]{algpseudocode}
\usepackage{caption}
\definecolor{hollywoodcerise}{rgb}{0.96, 0.0, 0.63}
\definecolor{lasallegreen}{rgb}{0.03, 0.47, 0.19}
\definecolor{hanpurple}{rgb}{0.32, 0.09, 0.98}
\definecolor{green(pigment)}{rgb}{0.0, 0.65, 0.31}
\definecolor{yellow}{rgb}{0.85, 0.85, 0.31}
\usepackage[pagebackref=true,breaklinks=true,letterpaper=true,colorlinks]{hyperref}
\hypersetup{colorlinks,linkcolor={red},citecolor={hollywoodcerise},urlcolor={red}}  

\usepackage[capitalize]{cleveref}
\crefname{section}{Sec.}{Secs.}
\Crefname{section}{Section}{Sections}
\Crefname{table}{Table}{Tables}
\crefname{table}{Tab.}{Tabs.}
\crefname{figure}{Fig.}{Figs.}
\crefname{algorithm}{Alg.}{Algs.}

\iccvfinalcopy 

\def\iccvPaperID{9438} 

\ificcvfinal\fi

\begin{document}

\title{Dynamic PlenOctree for Adaptive Sampling Refinement in Explicit NeRF}

\author{Haotian Bai$^{1}$
\quad
Yiqi Lin$^{1}$
\quad
Yize Chen$^{1, 2}$$^{*}$
\quad
Lin Wang$^{1,2}$\thanks{L. Wang and Y. Chen are the corresponding authors.}
\and
\affmark[1] VLIS LAB, AI Thrust, HKUST(GZ)\quad
\affmark[2] Dept. of CSE, HKUST\\
\quad
{\tt\small haotianwhite@outlook.com, linyq29@gmail.com, yizechen@ust.hk, linwang@ust.hk} \\
{\small Project homepage: \url{https://vlislab22.github.io/DOT/}}
}


\maketitle
\ificcvfinal\thispagestyle{empty}\fi

\begin{abstract}
The explicit neural radiance field (NeRF) has gained considerable interest for its efficient training and fast inference capabilities, making it a promising direction such as virtual reality and gaming. In particular, PlenOctree (POT)~\cite{yu2021plenoctrees}, an explicit hierarchical multi-scale octree representation, has emerged as a structural and influential framework.
However, POT's fixed structure for direct optimization is sub-optimal as the scene complexity evolves continuously with updates to cached color and density, necessitating refining the sampling distribution to capture signal complexity accordingly.
To address this issue, we propose the dynamic PlenOctree (\textbf{DOT}), which adaptively refines the sample distribution to adjust to changing scene complexity.
%
Specifically, DOT proposes a concise yet novel hierarchical feature fusion strategy during the iterative rendering process.
Firstly, it identifies the regions of interest through training signals to ensure adaptive and efficient refinement. 
Next, rather than directly filtering out valueless nodes, DOT introduces the sampling and pruning operations for octrees to aggregate features, enabling rapid parameter learning. 
Compared with POT, our DOT outperforms it by enhancing visual quality, reducing over $55.15$/$68.84\%$ parameters, and providing 1.7/1.9 times FPS for NeRF-synthetic and Tanks $\&$ Temples, respectively.  
\end{abstract}

\maketitle


\begin{figure}[t!]
\centering

\includegraphics[width=\linewidth]{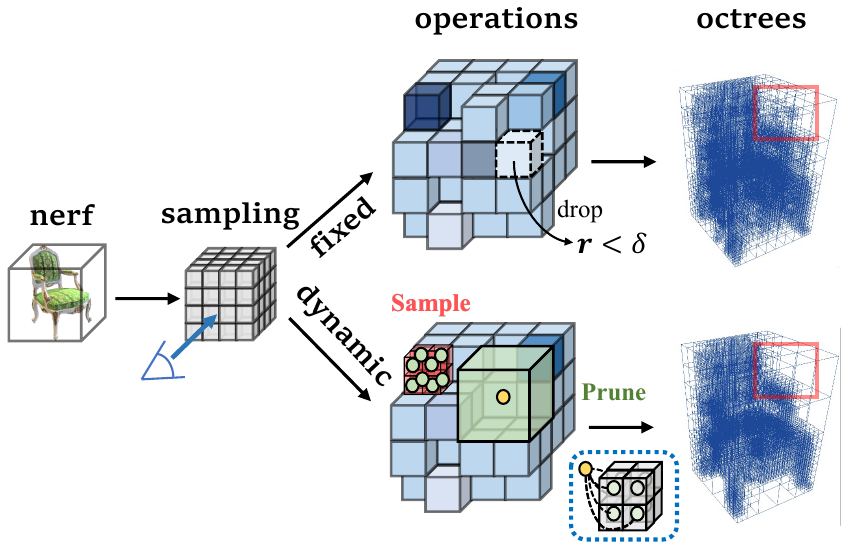}
\caption{
%
While the POT framework is effective, its fixed octree structure can limit its adaptability to varying scene complexities. We introduce hierarchical feature fusion with sampling/pruning to overcome this limitation, as illustrated by the dashed box below. Varying colors on the grid represent the training signals. Internal nodes are denoted in \textbf{\textcolor{orange}{orange}}, while leaf nodes in \textbf{\textcolor{green(pigment)}{green}}. Pruning occurs in regions of weak signal, where cached properties in leaf nodes are aggregated, and the averaged value is propagated to internal nodes, which become the new leaves. Complementary sampling takes place in the red regions. The resulting sampling distribution exhibits significant improvement, as highlighted by the red boxes in our final octree results.
}
\label{fig:teaser}

\end{figure}

\section{Introduction}
\label{sec:intro}
Rendering photo-realistic scenes and objects is crucial for providing users with an immersive and interactive experience in virtual reality~\cite{gao2022nerf,zhang2022neuvv} and metaverse~\cite{lee2021all,peng2021animatable,zhou2022vetaverse}.
Neural radiance field (NeRF)~\cite{nerf} has emerged as a promising solution for modeling 3D scenes and objects with only calibrated multi-view images.
Many subsequent approaches~\cite{mip-nerf, lindell2021autoint, muller2022instant,ref-nerf,zhang2022ray} have been proposed to further enhance NeRF's rendering power in terms of the training time, inference speed, and quality. 
PlenOctree (POT)~\cite{yu2021plenoctrees} stands out among these approaches. It employs an explicit octree structure with spherical basis functions to accelerate and enhance the rendering quality. Such method achieves high-quality rendering at over 150 FPS on an NVIDIA V100 GPU, opening up new possibilities for real-time and high-quality rendering, utilizing explicit octrees.
Moreover, POT bridges the implicit and explicit NeRFs. Specifically, it demonstrates that POT can transform the implicit NeRFs into the octree representation, further boosting the NeRF training by five orders of magnitudes with the early stop.

Technically, POT's main contributions can be classified into two folds: quality enhancement with non-Lambertian effects and a multi-scale sampling strategy with an octree.
Firstly, POT employs spherical harmonics~(SH) to model the non-Lambertian view-dependent effects and directly stores them along with the density in POT's leaves for fast training and inference. 
Secondly, POT employs a multi-scale approach that pre-samples density and SH using a multi-level tabulated volume. 
The resulting octree structure facilitates capturing intricate features with deeper octree sub-divisions and diversifies the sampling density according to the distribution of signal complexity, \ie the expressiveness of cached color and density through the scene. 

However, after the octree construction, POT keeps the division fixed for optimization. We argue such a process is sub-optimal,  as the signal complexity can vary during training. 
Therefore, its initial sample distribution may not provide a sufficient sampling rate, leading to aliasing or oversampling issues. Our proposed dynamic design addresses emerging questions on how to calibrate the spatial distribution and aggregate learned features during its construction.
In addition, there has been growing interest in recent research for striking a balance between compactness and expressiveness in NeRF representation, with sampling methods falling into two main categories. The first category, known as importance sampling~\cite{kurz-adanerf2022, nerf}, involves predicting the locations of samples and allocating more resources to complex regions. 
Although sampling on regions of interest can effectively capture signal complexity, predicting the locations can be computationally expensive, especially when dealing with millions of rays.
The second category~\cite{Liu2020NeuralSV, plenoctrees, Yu2022PlenoxelsRF}, such as POT, relies on stratified dense sampling, followed by rejecting samples below a certain threshold.
While filtering saves computational resources by discarding valueless samples, the heuristic rejection process may accidentally drop valuable samples, potentially lowering performance. Moreover, this process can break the global view consistency as the volume rendering strives to  build a consistent 3D representation across all views.

In this paper, we propose a concise yet novel \textit{hierarchical feature fusion} approach that combines the benefits of importance sampling and rejection methods. 
Specifically, we exploit the evolving signal complexity during training by utilizing the ray weight or density values as training signals through importance sampling, which incurs no additional cost. 
With these guided signals, we employ a rejection method to prune valueless regions and selectively sample the most promising regions to capture fine details, striking a balance between accuracy and efficiency.
We temporarily fix the octree and optimize the cached properties to adapt to recent modification. The entire process is iterative, allowing us to progressively calibrate the octree structure to increase its compactness and capture more details as training progresses.

Notably, \textit{we do not directly drop out voxels but instead, fuse learned features while modifying the octree division}. As depicted in \cref{fig:teaser}, our novel hierarchical feature fusion approach facilitates adaptive refinement and enables rapid parameter learning through octree sampling and pruning operations. We demonstrate its effectiveness by showing that it can save around 60\% parameters while enhancing rendering quality across different scenes.

Our extensive experiments demonstrate the benefits of DOT over POT. Our method enriches rendering views, reduces the number of required parameters by around 60\%, and nearly doubles the rendering speed. 
Furthermore, we offer more control over sampling and pruning strength, which enhances flexibility when working with scenes of varying complexity.
Specifically, 
DOT achieves excellent performance, allowing for rendering an 800x800 image at $452$ FPS on an RTX 3090 GPU, achieving 1.8 times the FPS of the POT model.

\begin{figure*}[t]
    \centering
    \includegraphics[width=\linewidth]{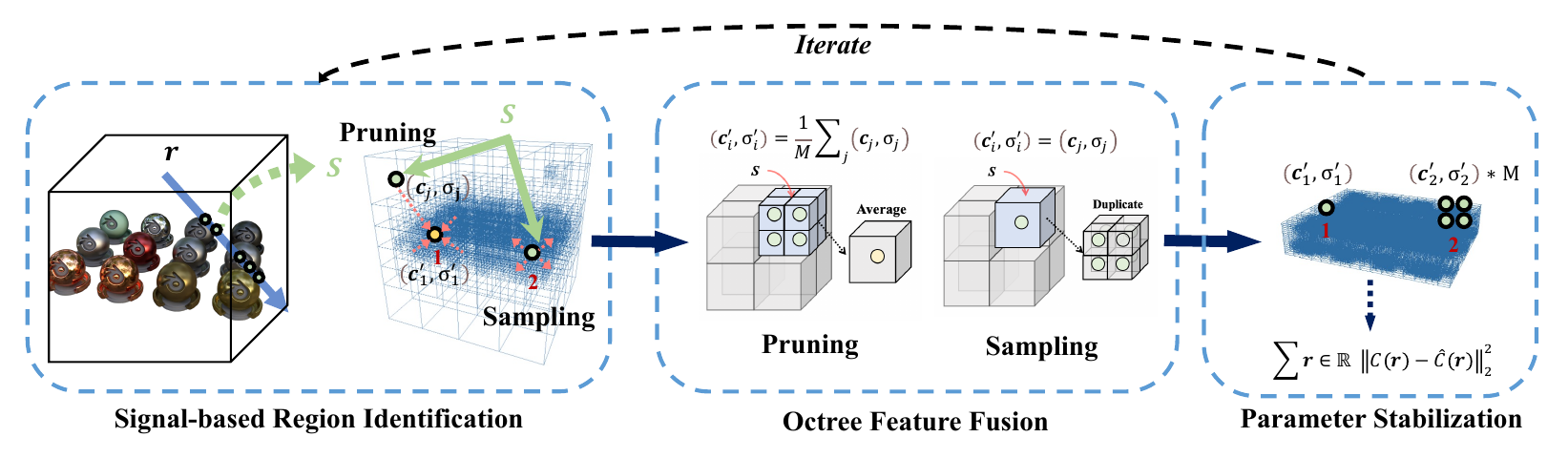}
    \caption{\textbf{Overview of DOT pipeline.} 
    %
The exact grid partitions are \textbf{\textcolor{green(pigment)}{green}} \textbf{leaf} nodes, while the \textbf{\textcolor{orange}{orange}} \textbf{internal} nodes are the parent nodes of leaves, only used for octree organization instead of physical space allocation.
Guided by the training signal $\mathbf{S}$,  DOT prunes valueless regions and aggregates their features into the internal nodes ``1". It also samples complex areas, such as the leaf nodes ``2", by propagating its learned properties into the newly allocated leaf nodes in the next level. 
    }
    \label{fig:framework}
\end{figure*}

In summary, this paper makes three major contributions:
\begin{itemize}
    \item We improve the fixed octree design in POT, allowing for iterative refinement of the octree structure based on training signals iteratively without introducing additional cost.
    \item We introduce the hierarchical feature fusion strategy to support the adaptive refinement of the octree division, enabling rapid parameter learning by aggregating features via octree sampling/pruning operations.
    \item Experiments on two benchmark datasets show that DOT can dramatically slim POT and accelerate the rendering speed while improving rendering quality.
\end{itemize}


\section{Related Work}

\noindent\textbf{Neural Radiance Fields (NeRF).}
NeRF~\cite{nerf} uses volume rendering to train coordinate-based MLPs that can directly predict color and opacity based on 3D position and 2D viewing direction. The resulting synthesized views have photo-realistic quality, and the differential volume rendering technique has been widely adopted in various applications, including scene and object relighting~\cite{srinivasan2021nerv,zhang2021nerfactor}, unbounded scenes~\cite{KaiZhang2020NeRFAA,barron2022mip}, dynamic scenes from videos~\cite{gao2021dynamic,pumarola2021d,xian2021space,tretschk2021non}, editable scenes, avatars~\cite{liu2021neural,yang2021learning}, and object surface reconstruction~\cite{azinovic2022neural,wang2021neus}. Recent advances in NeRF focus on improving training speed~\cite{lindell2021autoint,muller2022instant,Liu2020NeuralSV} and rendering quality~\cite{mip-nerf,ref-nerf,KaiZhang2020NeRFAA}. In particular, POT aims to provide a real-time high-fidelity rendering experience, as taking minutes to render a novel view is unsuitable for real-time rendering or 3D interaction.
Existing frameworks for boosting rendering speed can be divided into three groups. The first group focuses on designing sampling-friendly models for fast rendering~\cite{rebain2021derf,garbin2021fastnerf,reiser2021kilonerf,neff2021donerf,lindell2021autoint,kurz-adanerf2022}. 
Secondly, \cite{hedman2021baking,Hedman2021BakingNR,Liu2020NeuralSV,reiser2021kilonerf} provide instant rendering by baking or caching weights into explicit spatial data structures from well-trained NeRF models. 
The other group of methods~\cite{chen2022mobilenerf, Yariv2023BakedSDFMN} represents NeRF using GPU-friendly meshes, boosting the speed by the standard polygon rasterization pipelines and rendering pipelines across different devices.

\noindent\textbf{Sampling Refinement.}
Dense sampling is usually necessary for pre-sampling the locations along rays to avoid unnecessary computation on sparse regions. Implicit NeRF \cite{nerf} uses stratified sampling followed by a second network to sample important regions based on previous prediction, while AdaNeRF \cite{kurz-adanerf2022} proposes the dual sampling network to disentangle samples based on sparsity. However, these methods increase inference and training complexity and bring massive costs regarding the millions of rays to render. Recent works optimize this sampling procedure by filtering unnecessary samples with threshold values to avoid additional queries. NSVF \cite{Liu2020NeuralSV} rejects points from stratified sampling to accelerate rendering, while \cite{plenoctrees, Yu2022PlenoxelsRF} filter unnecessary samples based on their rendering contribution after dense sampling. However, the rejection method may accidentally drop valuable samples, potentially lowing performance and breaking the global view consistency as the volume rendering strives to
build a consistent 3D representation across all views.
Another approach\cite{muller2022instant, Chen2022TensoRFTR, Fang2022FastDR} represented by Instant-NGP\cite{muller2022instant}, uses multi-resolution grids that leverage meaningful learning. This method samples the locations that cause significant changes based on the magnitude of gradients in a fixed grid space. However, meaningful learning is not applicable in dynamic scenarios, as the gradients disappear when  division is eliminated.

\noindent\textbf{Explicit Spatial Structure in NeRF.}
Storing learnable features in grid structures is a promising alternative to MLPs for fast rendering because cached values can be accessed directly. Recent research has explored various explicit representations for NeRF, including dense grids~\cite{karnewar2022relu,sun2022direct}, sparse 3D voxel grids~\cite{Liu2020NeuralSV,Yu2022PlenoxelsRF, yu2021plenoctrees}, multiple compact low-rank tensor components~\cite{chen2022tensorf}, 3D point clouds~\cite{xu2022point}, and multi-resolution hash maps~\cite{muller2022instant}. The octree structure has received increasing attention in NeRF due to its multi-scale space division, which allocates more samples to complex regions while quickly skipping sparse regions. ACORN\cite{martel2021acorn} introduced a hybrid implicit-explicit network using octree decomposition and an adaptive resource allocation strategy based on signal complexity.
DOT builds on the adaptive sampling idea that uses the training signals to guide the calibration, but we focus more on fusing features in the octree while modifying its structure. This is necessary as features in explicit representations are unable to adapt to the changes as effectively as implicit representations.
\section{Method}
\label{sec:Method}
\noindent \textbf{Overview:}
 \label{overview}
DOT introduces a hierarchical feature fusion method designed for octrees. The method comprises three main components: signal-based region identification (see \cref{sec:signal_id}), octree feature fusion (see \cref{sec:adaptive_op}), and parameter stabilization (see \cref{sec:stab}). The DOT pipeline, as shown in \cref{fig:framework}, involves an iterative and adaptive modification process based on evolving training signals. Firstly, the method identifies the regions of interest by tracking the instant training signals during the rendering process. Next, guided by the signal, DOT adaptively modifies those regions by pruning and sampling operations, aggregating the features across the octree division. Finally, after stabilizing the learned parameters, DOT iteratively calibrates the octree to capture more details and compact the representation.

\noindent \textbf{Preliminaries:}
 \label{Preliminaries}
%
The volume rendering process of POT is fully differentiable, allowing for direct optimization of cached properties such as spherical harmonics (SH) and opacity within the octree structure. Moreover, POT determines the ray-voxel intersections for each ray in the octree structure, producing a sequence of unevenly distributed sampling segments $\{{\delta_i}\}_{i=1}^{N_\mathbf{r}}$ that adaptively fit the initial signal complexity by skipping sparse voxels in one step while not missing the small ones. 
The rendering process, based on the classic work by Kajiya \etal~\cite{JamesTKajiya1984RayTV}, infers the color of a ray $\hat{C}(\textbf{r})$ by integrating $N_\mathbf{r}$ samples along the ray. Specifically, the pixel's predicted color $\hat{C}(\textbf{r})$ is approximated by accumulating the colors of the samples weighted by $Q_i$:
\begin{align}
\label{eq:vol_rend}
\hat{C}(\textbf{r}) &= \sum_{i=1}^{N'}T_iQ_i\textbf{c}_i,\\
Q_i &= 1-\exp(-\sigma_i\delta_i),
\label{eq:Q_def}   
\end{align}
%
where the light transmitted through a batch of rays $\mathbf{r}$ to sample $i$ is represented by $T_i = \exp(-\sum_{j=0}^{i-1}\sigma_j\delta_j)$; $\sigma_i$ denotes the opacity of the sample. $ Q_i$ denotes the light contribution of sample $i$, and $\mathbf{c}_i$ is the color vector in the form of SH.
$\hat{C}(\textbf{r})$ is optimized to approximate the ground truth color $C(\textbf{r})$ by minimizing MSE loss $\sum_{\textbf{r}\in\mathbb{R}}\|C(\textbf{r})-\hat{C}(\textbf{r})\|_2^2$ over the set of rays $\mathbb{R}$ in the training images.

%

\subsection{Signal-based Region Identification}
\label{sec:signal_id}
The accuracy of the predicted color vector $\hat{C}(\textbf{r})$ in \cref{eq:vol_rend} depends on the quality of ${\{\delta_i\}_{i=1}^{N_\mathbf{r}}}$, \ie the sampling density to reconstruct $C(\textbf{r})$ to catch up the changes in scene complexity.
Recent studies \cite{Xing2022MVSPlenOctreeFA, Yu2022PlenoxelsRF, Hedman2021BakingNR} have highlighted the significance of the signal response such as $\boldsymbol{\sigma}$ and $\boldsymbol{Q}$ in \cref{eq:Q_def} for regularization when constructing NeRF representations.
In the case of POT, the octrees are converted from well-trained implicit NeRFs. Thus resampled values can be taken with high confidence. This allows DOT to leverage these properties directly to guide the latter modifications.
In practice, we utilize importance sampling, which prioritizes selecting samples from areas with a high signal response. Additionally, DOT compacts the structure by aggregating features from regions with weak signal values.

\begin{figure}[t!]
    \centering
    \includegraphics[width=\linewidth]{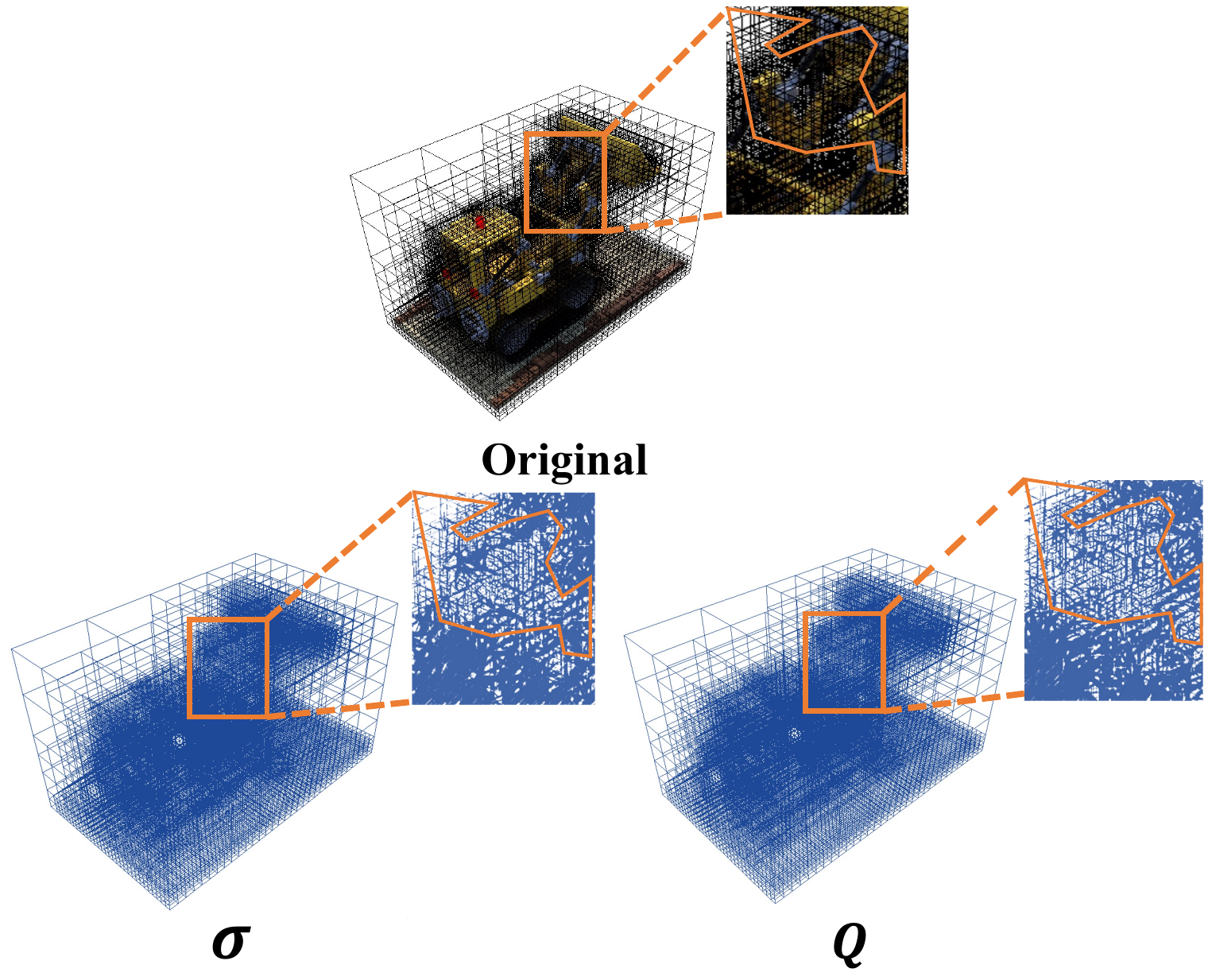}
    \caption{\textbf{Comparison on training signals. } 
    We highlight the difference in different choices of training signals, including the opacity $\boldsymbol{\sigma}$ or the ray-weight $\mathbf{Q}$ in the boxed regions, where pruning by $\mathbf{Q}$ eliminates more invisible voxels than $\boldsymbol{\sigma}$.
    We keep the similar PSNR($\pm0.05$) (\cref{tb:abls}) for a fair comparison. 
    \textcolor{red}{Please zoom in to see more details}. }
    \label{fig:prune_targets}
\end{figure}

Specifically, denote the total number of voxel samples as $N=\sum_{\mathbf{r}} N_\mathbf{r}$ with each ray interacts with $N_\mathbf{r}$ divisions. The ray weights $\textbf{Q}=\left \{\sum_\mathbf{r} Q_{i,\mathbf{r}} \right\}_{i=1}^N$ reflects the overall rendering contribution, and the density $\boldsymbol{\sigma} = \{\sigma_i\}_{i=1}^N$. 
Our prune-only experiment shown in \cref{fig:prune_targets} compares their effectiveness. It demonstrates that choosing $\textbf{Q}$ as the target training signal eliminates more invisible voxels than $\boldsymbol{\sigma}$, which is crucial for fast ray inference. For instance, the arm of the \emph{lego} model turns out to be more compact with $\textbf{Q}$. Further quantitative analysis in \cref{tb:abls} proves that the $\textbf{Q}$ approach is $20\%$ faster than $\boldsymbol{\sigma}$ despite sharing similar memory and fidelity.


%
%
%
%
%
%

\subsection{Octree Feature Fusion}
\label{sec:adaptive_op}
After identifying the region to modify based on the training signals, the next step is to determine how to aggregate the cached features during the modification process. A common approach is to discard the voxels with signal values below a fixed threshold.
However, simply discarding voxels based on a fixed threshold can lead to the loss of valuable learned features and view consistency in volume rendering, as the volume rendering(\cref{eq:vol_rend}) strives to
build a consistent 3D representation across all views.
Recent works on explicit NeRFs\cite{Yu2022PlenoxelsRF, Liu2020NeuralSV, plenoctrees} have shown the benefits of learning properties in a course-to-fine manner, incrementally adding details based on previous estimates. These methods typically begin with dense stratified sampling to learn the feature distribution and context.  However, during the transition, learned features are removed, resulting in fewer parameters at the expense of losing globally consistent features.

\begin{figure}[t]
    \centering
    \includegraphics[width=0.9\linewidth]{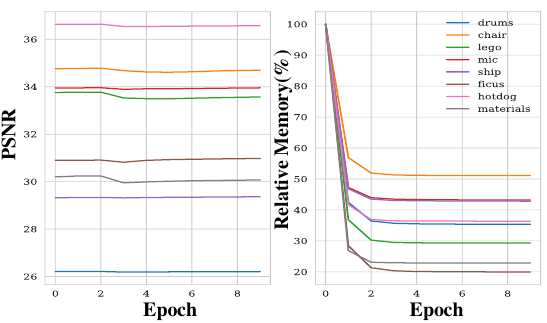}
    \caption{\textbf{The training progress with only pruning operation. }Denote the relative memory as the model size DOT$/$PlenOctree. We train the prune-only model for ten epochs using the signal $\mathbf{Q}$ on the NeRF-synthetic dataset. Surprisingly, DOT's single pruning operation improves the quality in \emph{ficus} while slimming over $80\%$ of memory. 
    }
    \label{fig:abl_prune_only}
\end{figure}

Our proposed solution for this issue is to use feature fusion to calibrate the octree structure. We base this approach on the neighboring assumption that features can be propagated across different levels of the octree structure,
\begin{equation}
\label{eq:pruning}
(\mathbf{c}_i', \sigma_i')=\frac{1}{M}\sum_j(\mathbf{c}_j, \sigma_j),
\end{equation}
where $M=8$ is the degree of the octree. 
\cref{eq:pruning} forms the basis of the octree pruning operation. As illustrated in \cref{fig:framework}, the internal nodes $\left \{(\mathbf{c_i}', \sigma_i')\right \}_i$ collect their leaves' features to represent the larger physical division. Consequently, this approach retains features and their global consistency learned from the volume rendering.
Besides, the complementary sampling operation can be denoted as, 
\begin{equation}
  (\mathbf{c}_j', \sigma_j')=(\mathbf{c}_i, \sigma_i),
  \label{eq:sample}
\end{equation}
which increases the sampling density in the selected partitions $\left\{(\mathbf{c}_j', \sigma_j')\right\}_j$ and turns $\left\{(\mathbf{c}_i, \sigma_i)\right\}_i$ into internal nodes. 
The effectiveness of our proposed approach is demonstrated in the prune-only experiment(see \cref{fig:abl_prune_only}), where the pruning operation significantly reduces the number of parameters without compromising the quality. This result underscores the importance of adapting the sample distribution in response to changes in signal complexity within the scene. In other words, a fixed sample distribution can result in suboptimal quality and high memory costs.

In addition, we introduce two hyperparameters $\tau$ and $\gamma$ to adjust the strength of pruning and sampling, thereby increasing the flexibility of the signal guidance across different scenes. 
The selected samples are defined based on signal response with thresholds as $\{i | Q_i\le \tau\}$ and $\{i | Q_i>\gamma\}$ for a trade-off between memory and performance. We observe that the proposed method provides better control over the level of detail. 
Specifically, \cref{fig:abls}(c) tells the \emph{ship} model pruned with $\tau=10$ takes surprisingly $3\%$ of memory required by POT while still preserving the necessary details.
To achieve more intensive pruning for sparse scenes, we introduce a recursive pruning option that enables marching into the upper part of the octree hierarchy. As shown in \cref{fig:abls}(a), the recursive pruning approach is more memory-efficient than the regular one-time pruning while causing a negligible degradation in rendering quality. Specifically, the recursive pruning approach merges nodes whose signal response values are below or equal to the threshold $\tau$ in a recursive manner, allowing for more excellent compression of the octree structure. The algorithmic details of the pipeline are presented in \textit{suppl.} materials.

\begin{table}[t]
\centering
\scalebox{0.68}{
\begin{tabular}{lcccccc} 
\toprule[1pt]
 & GPU & Memory$\downarrow$  & PSNR$\uparrow$ & SSIM$\uparrow$& LPIPS$\downarrow$  & FPS$\uparrow$\\
 \hline
 
Neural Volumes\cite{NeuralVolumes}    &Tesla V100     & -  & 23.70     &0.834    & 0.260  & 1.0        \\
NSVF\cite{liu2020neural}   &Tesla V100    & - &31.75      &0.953    &\textbf{0.047}   &0.8         \\
AutoInt(8)\cite{lindell2021autoint}  &Tesla V100  & -     &25.55       &0.911    &0.170   &0.4         \\
Plenoxels\cite{Yu2022PlenoxelsRF} & - & - &31.71 &0.958 &\underline{0.049} &$\sim15.0$\\
FastNeRF\cite{garbin2021fastnerf} & RTX 3090   &- &29.97 &0.941    &0.053    & 238.1 \\
SNeRG\cite{hedman2021baking} & RTX2080Ti & 2.71 &30.38  &0.950    & 0.050    & 207.26 \\
MobileNeRF~\cite{chen2022mobilenerf} &RTX2080Ti   &\textbf{0.54} &30.90      &0.947    &0.062    &\textbf{744.91}   \\
PlenOctree~\cite{yu2021plenoctrees} &RTX3090   &1.93 &\underline{31.71}      &0.958      &0.053    &250.1    \\
\hline
Ours   &RTX3090 & 0.87     &\textbf{32.11 }    &\textbf{0.959}    &0.053    &452.1      \\
Ours(R) &RTX3090  & \underline{0.80}     &\underline{32.03}      &\underline{0.958}   &0.054    &\underline{474.2}      \\
\bottomrule[1pt]
\end{tabular}
}
\caption{
\textbf{Quantitative results on the NeRF-synthetic.} The memory is measured by GB. The memory is measured by GB. The \textbf{best} and the \underline{second-best} results are highlighted. (R) denotes the recursive pruning.
}
\label{tb:SYN_perf}
\end{table}
\begin{table}[!t]
\centering
\scalebox{0.68}{
\begin{tabular}{lcccccc} 
\toprule[1pt]
 & GPU & Memory$\downarrow$  & PSNR$\uparrow$ & SSIM$\uparrow$& LPIPS$\downarrow$  & FPS$\uparrow$\\
 \hline
 Neural Volumes\cite{NeuralVolumes}  &Tesla V100  & -       &23.70      &0.834    &0.260   &1.0         \\
 NSVF\cite{liu2020neural}  &Tesla V100 & -    &28.40      &0.900    &0.153   &0.2         \\
PlenOctree~\cite{yu2021plenoctrees}  &RTX3090 &3.53      &28.00      & 0.917   &0.131    &74.0       \\
\hline
Ours   &RTX3090 &1.10   &28.28      &0.922   &0.121    &186.2      \\
Ours(R) &RTX3090  &0.92    &28.25      &0.922   &0.122    &216.1      \\
\bottomrule[1pt]
\end{tabular}
}
\caption{
\textbf{Quantitative results on the Tanks $\&$ Temples.} 
}
\label{tb:TT_perf}
\end{table}

\begin{table}
\centering
\scalebox{0.8}{
\begin{tabular}{lccccc} 
\hline
Methods & Memory(GB)$\downarrow$ & PSNR$\uparrow$ & SSIM$\uparrow$ & LPIPS$\downarrow$  & FPS$\uparrow$\\
 \hline
$Sm,Pr$ & 0.815 & 32.113 &0.959 &0.053 &531.144 \\ 
 \hline
$Pr_\sigma$  &0.611      &31.753         &0.9578   &0.054  &419.686        \\
$Pr_Q$    &0.600  &31.748    &0.9575        &0.055 &569.288       \\
\hline
\end{tabular}
}
\caption{
\textbf{Ablations on NeRF-synthetic.} 
We experiment on NVIDIA A100 to evaluate the effectiveness of joint sampling and pruning(with $\boldsymbol{Q}$) (referred to as $Sm, Pr$) and individual pruning operations targeting different training signals.
Specifically, we used $Pr_\sigma$ to denote pruning based on $\boldsymbol{\sigma}$ and $Pr_Q$ on $\boldsymbol{Q}$.
}
\label{tb:abls}

\end{table}

\begin{figure*}[t!]
    \centering
    \includegraphics[width=\linewidth]{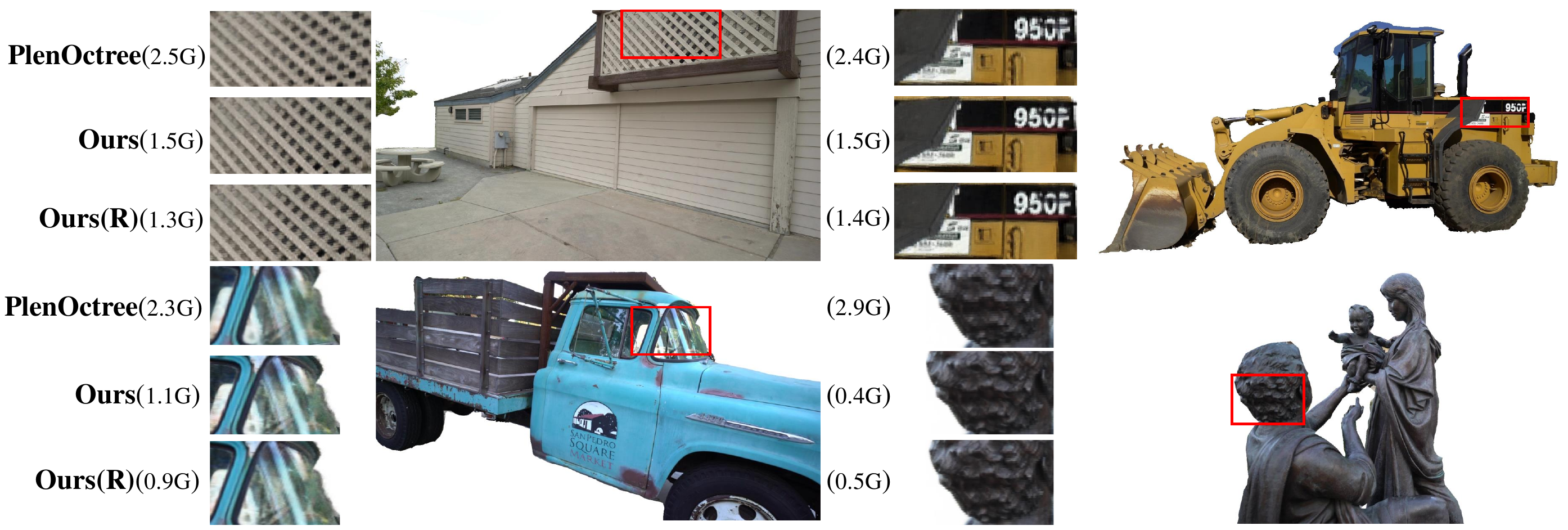}
    \caption{\textbf{Tanks $\&$ Temples qualitative results.}
    Interestingly, our methods surpass POT in quality with over half parameter size shrinkage, adding more details on the high-frequency regions. \textcolor{red}{Please zoom in to see more details}.
    }
    \label{fig:tt_data_fig}
\end{figure*}

\subsection{Parameter Stabilization}
\label{sec:stab}
We regularly fix the octree structure for $T$ epochs to allow the model to stabilize the cached properties. Since the structure is revised by pruning and sampling operations, it takes time for the cached values to adapt to the calibrated octree division and update the scene complexity accordingly.
To optimize the high-dimensional voxel coefficients, we adopt RMSProp\cite{hinton_2018} as the optimization algorithm, as the non-convexity of the rendering formula \cref{eq:vol_rend} makes direct optimization challenging.
%
The steady increase in PSNR illustrated in \cref{fig:abls}(a) proves the effectiveness. 





\section{Experiments}

\subsection{Experiment Setup}
\textbf{Dataset}: We used two datasets in our experiments. The first is the NeRF-synthetic dataset, which consists of eight scenes with single objects, including \emph{hotdog}, \emph{materials}, \emph{ficus}, \emph{lego}, \emph{mic}, \emph{drums}, \emph{chair}, and \emph{ship}. The views are at a resolution of $800\times800$, with 100 views for training and 200 for testing per scene. The second dataset is a subset of Tanks \& Temples from NSVF~\cite{Liu2020NeuralSV}. It contains five scenes of real objects captured by an inward-facing camera that circles the scene, and we used foreground masks provided by NSVF. Each scene contains 152-384 images with a resolution of $1920\times1080$.

\textbf{Baselines}: DOT is built based on the same pretrained NeRF-SH models as POT, where all trained representations store density and SH coefficients converted from NeRF-SH. The grid size is set to $512^3$, and the pre-trained models use 16 and 4 SH components for the synthetic and Tanks $\&$ Temples datasets, respectively.

\textbf{Implementation details}: 
We use stochastic image samples to evaluate MSE loss. 
Specifically, 
the training pipeline takes 100 epochs, with an interval of $T=20$. 
The learning rate is fixed in the main experiments, which is set to $0.1$ and $0.01$ for $\sigma$ and $\mathbf{c}$.
Furthermore, 
%
We apply $\tau = 1, 10$ for the synthetic and the Tanks $\&$ Temples datasets, respectively, and $\gamma$ is set to 0.01 for both. 

\textbf{Compression}:
To compare the effectiveness of compressed octrees, we use the same median-cut algorithm\cite{HeckbertPaul1982ColorIQ} to quantize the SH coefficients. Our compressed DOT models also support in-browser rendering using WebGL. We directly squeeze the SH-16 POT models for a fair comparison.

\begin{figure*}[!t]
    \centering
    \includegraphics[width=\linewidth]{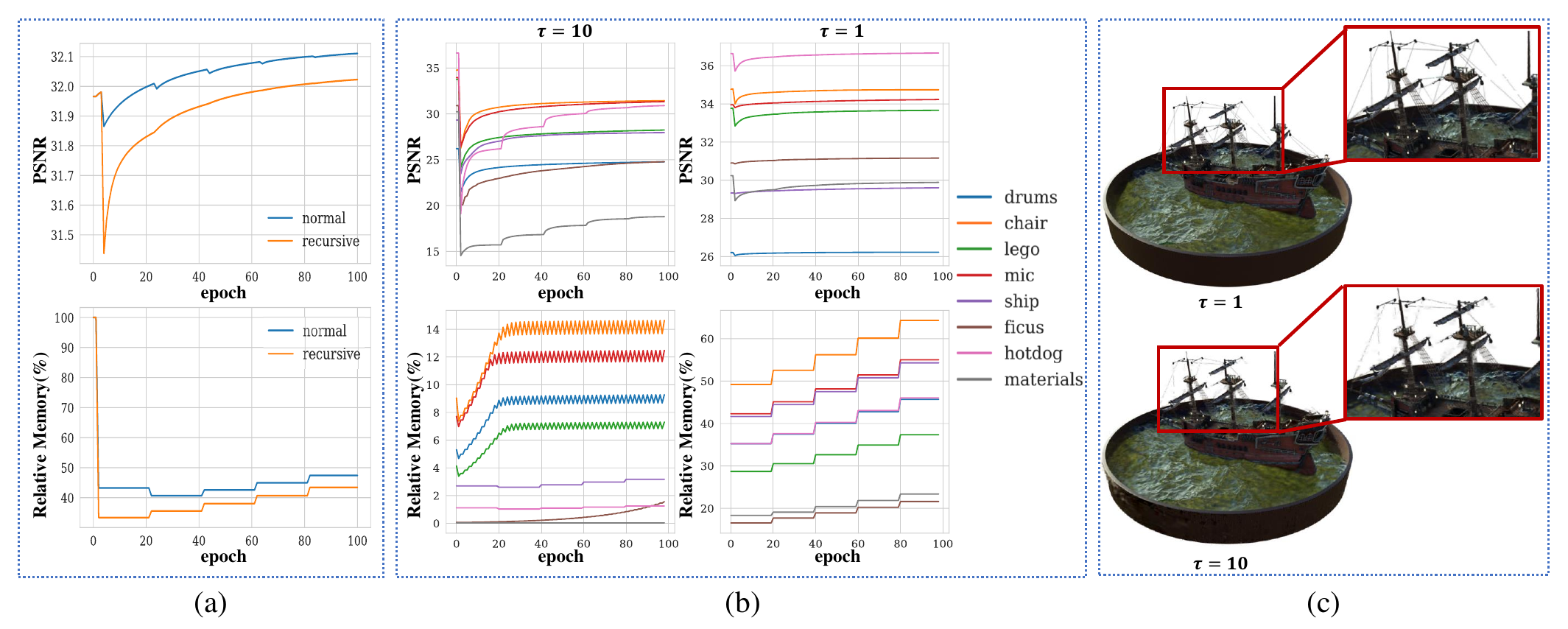}
    \caption{\textbf{The ablation studies
    }
    Denote the relative memory as the percentage of DOT's memory cost compared to POT, \ie DOT/POT. 
    (a) The comparison between the recursive pruning and the one-time pruning(normal) with the average PSNR and the relative memory usage in NeRF-Synthetic test scenes. We train both models using the same sampling algorithm with different pruning options on the NeRF-Synthetic for 100 epochs. 
    (b) Using a similar experiment setting as (a), we evaluate the effect of pruning strength $\tau$. Specifically, we alter the pruning strength by varying the threshold $\tau=1$ and $\tau=10$ to test its effect.  
    (c) The qualitative comparison for (b) on the \emph{ship} scene in the synthetic dataset.  \textcolor{red}{Please zoom in to see more details}.
    }
    \label{fig:abls}
\end{figure*}
\subsection{Main Results}
We conduct a comprehensive evaluation of DOT compared with POT on both synthetic and real datasets, presented in \cref{tb:SYN_perf} and \cref{tb:TT_perf} for the NeRF-synthetic and Tanks $\&$ Temples datasets, respectively.
Our method achieves significant parameter savings compared to the original POT models, particularly with over $70\%$ memory savings on Tanks $\&$ Temples. Moreover, despite having fewer parameters, our method outperforms it in all metrics, including PSNR, SSIM, and LPIPS~\cite{zhang2018unreasonable}.
The superior performance of DOT can be attributed to its ability to refine the sample distribution to the varied signal complexity dynamically. As demonstrated by \cref{fig:tt_data_fig}, DOT provides more details in complex regions, such as sharper reflections on \emph{windows} and more evident edges on \emph{fences}.
Additionally, \cref{fig:data_fig} shows the more compact structure of DOT, resulting in fewer ray intersections, explaining our significant rendering speed boost.
Specifically, in the \emph{materials} scene, the sparse space collapses into a dense box that tightly fits the metal balls while enhancing the density on their surface. 

Furthermore, we compare DOT with other state-of-the-art methods, including 
Neural Volumes\cite{NeuralVolumes}, 
NSVF\cite{Liu2020NeuralSV}, Plenoxels\cite{Yu2022PlenoxelsRF}, AutoInt\cite{lindell2021autoint}, FastNeRF\cite{Fang2022FastDR}, SNeRG\cite{hedman2021baking}, and MobileNeRF\cite{chen2022mobilenerf}. DOT achieves state-of-the-art rendering quality and stands out regarding memory usage and rendering speed. Compared to MobileNeRF with polygon rasterization, DOT provides more visually appealing rendering results. Furthermore, the gradient-based mesh representations in MobileNeRF may be better suited to certain applications, with the potential crash for inaccurate surfaces or illumination effects. Therefore, it is a trade-off to select between DOT and MobileNeRF for different purposes.


\subsection{Ablation Study}
In this section, we conduct ablation studies to investigate the design choices of our proposed DOT in detail.

\noindent \textbf{Progressive calibration.}
We argue that naive rejection by heuristic threshold without further feature fusion leads to suboptimal quality and excessive memory usage. 
We verify it through the study presented in \cref{fig:abl_prune_only}, which compares the rejection-based POT and feature-fusion-based DOT.
The study shows that when only pruning is applied, the PSNR of DOT remains competitive or even outperforms POT after over 60$\%$ reduction in memory. 
Moreover, it also suggests that memory reduction is more intensive in the first few epochs and gradually stabilizes.
This could imply that DOT's sample distribution approaches the optimal sampling rate with progressive calibration.
We also discover that, 
the scene like \emph{ficus} can reduce more than 80$\%$ parameters while improving rendering quality.
Thus, it is not ideal to reject samples heuristically, and DOT makes a difference. 

\noindent \textbf{Pruning target.}
We demonstrated in \cref{fig:prune_targets} that adopting $\boldsymbol{Q}$ as the signal target instead of $\boldsymbol{\sigma}$ yields better results. Here, we provide a quantitative analysis of the outcome. We performed prune-only on both targets for ten epochs, and the training progress for the target $\mathbf{Q}$ is shown in \cref{fig:abl_prune_only}. Table \ref{tb:abls} shows that despite similar memory reduction and quality improvement between $\sigma$ and $\mathbf{Q}$, our results show that the $\boldsymbol{Q}$ approach has a considerable FPS gain of around $20\%$ while maintaining similar quality compared to its counterpart.

\begin{figure*}[t]
    \centering
    \includegraphics[width=\linewidth]{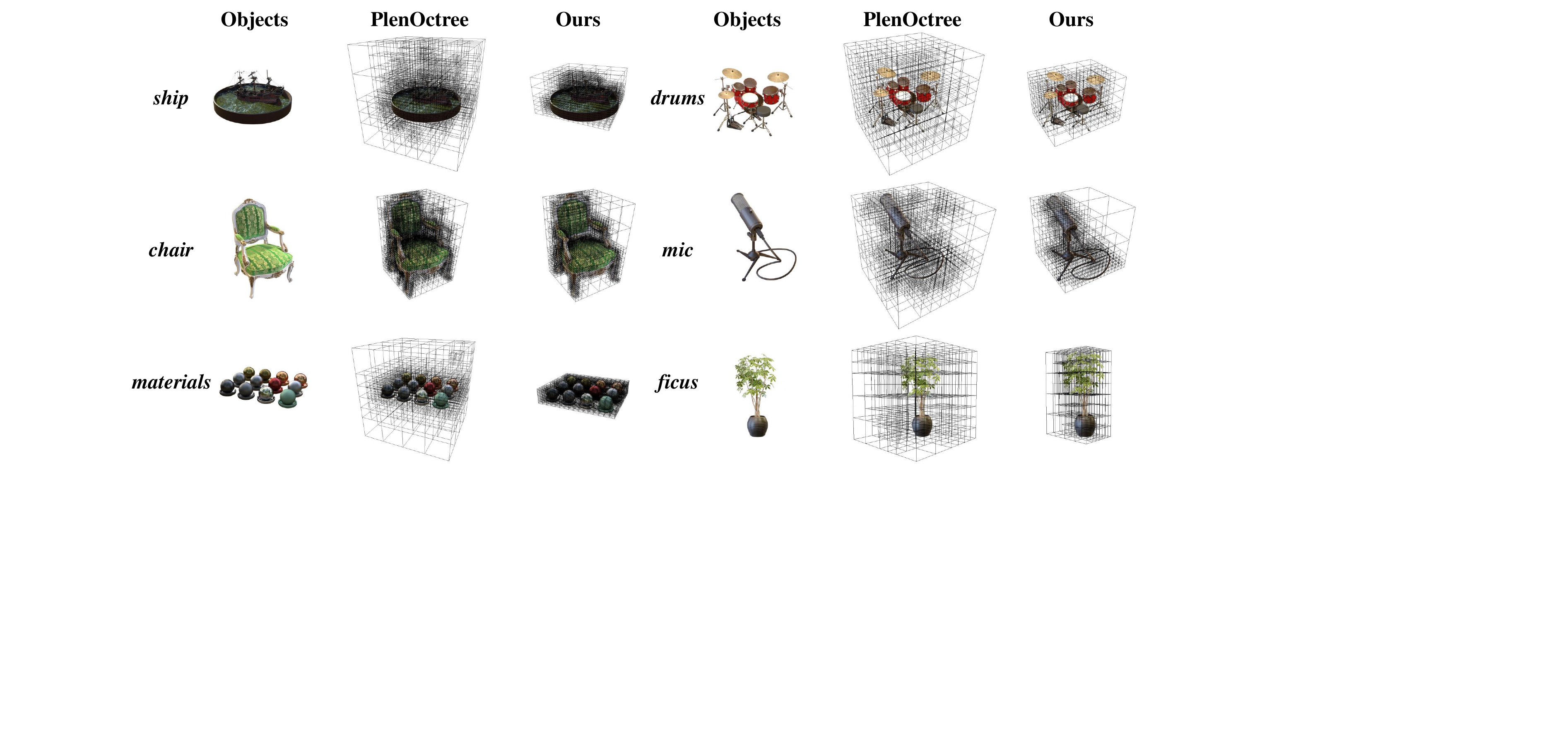}
    \caption{\textbf{NeRF-synthetic qualitative results}. We find that our methods produce more compact Octree structures.}
    \label{fig:data_fig}
\end{figure*}

\noindent \textbf{Pruning strength. }
We discuss the option for using recursive pruning  to eliminate the unnecessary nodes more thoroughly. 
%
%
As depicted in \cref{fig:abls}(a), recursive pruning brings approximately 5-8$\%$ more memory reduction but leads to a degradation of about 0.1 PSNR. Therefore, the recursive option is more beneficial for large scenes with many sparse regions, such as \emph{ficus} or \emph{materials}, as shown in \cref{fig:data_fig}.
On the other hand, we introduce $\tau$ to adjust the strength of pruning to enhance DOT's adaptiveness. To evaluate its effect, we set $\tau$ to 1 and 10 and test both memory reduction and PSNR on the NeRF-synthetic dataset. We observe that $\tau=10$ maintains the major context with only about $3\%$ of POT's parameters, indicating its flexibility to fit different demand levels. However, as shown in \cref{fig:abls}(b), higher $\tau$ values may set a bottleneck for potential quality enrichment, especially for complex scenes \eg, \emph{materials} and \emph{ficus}. For instance, we observe that memory frequently jitters, which may indicate that the key voxels in rendering are dropped, and further sampling operations may struggle to recover the original quality. Therefore, choosing a proper $\tau$ is necessary to balance the trade-off between quality and model size.

\begin{table*}[t!]

\centering
\setlength{\tabcolsep}{8mm}

\scalebox{0.8}{





\begin{tabular}{c|ccc|ccc} 
\hline
tt$/$syn & \multicolumn{3}{c|}{Full Models} & \multicolumn{3}{c}{Compressed~Models} \\ 
\hline
Method & PlenOctree & Ours & Ours(R) & PlenOctree & Ours & Ours(R) \\ 
\hline
Memory(GB) & 2.62$/$1.94 & 0.82$/$1.1 & 0.79$/$0.92 & 0.65$/$0.39 & 0.26$/$0.20 & 0.23$/$0.19 \\ 
\hline
FPS (A100) & 91$/$326 & 177$/$487 & 202$/$518 & 113$/$329 & 183$/$531 & 208$/$565 \\
FPS (RTX3090) & 74$/$250 & 186$/$452 & 216$/$474 & 109$/$289 & 270$/$467 & 239$/$492 \\
FPS (MX150) & \XSolidBrush $/$\XSolidBrush & \XSolidBrush$/$11 & \XSolidBrush$/$12 & \XSolidBrush$/$\XSolidBrush & 3$/$12 & 3$/$13 \\
\hline
\end{tabular}
}
\caption{
\textbf{The cross-device test on memory, and FPS on NeRF-Synthetic(syn) and Tanks $\&$ Temples scenes(tt).}
Denote the abbreviations for NVIDIA devices including  \underline{A100 PCIE 40GB }as A100, \underline{GeForce RTX 3090 24GB} as RTX3090, and \underline{GeForce MX150 2GB} as MX150. 
\XSolidBrush denotes not available due to overflow, 
and each item is the performance on tt/syn.}

\label{tb:cs_devices}
\end{table*}

\noindent \textbf{Necessity of sampling. }
We propose a complementary sampling operation to enhance high-frequency details and remedy accidental errors introduced from pruning within the high signal response regions. To demonstrate the necessity of this operation, we conduct an experiment comparing the prune-only models and full models, as presented in Table \ref{tb:abls}. The results demonstrate that sampling improves the PSNR by 0.15dB at the cost of approximately 10$\%$ memory that is to be reduced, which is a considerable boost regarding the cost.
While pruning alone can improve rendering fidelity by removing both valueless and incorrect samples that do not fit the training signals, its drawback is its inability to increase the sampling rate to keep up with the signal complexity. 
For instance, \emph{material} scene on the NeRF-synthetic is a challenging case to render due to its highly varied surface reflection, which requires a dense sampling distribution around the metal balls' surface. 
However, as shown in Figure \cref{fig:abls}(b), the joint operation preserves the original quality of the scene as the training progresses, while the prune-only experiment in \cref{fig:abl_prune_only} shows a decreased PSNR after pruning. 
It tells that the high-frequency features of surface reflections (the parameter stabilization box) in \cref{fig:framework} may not be recovered properly without sampling.
Therefore, we assert that sampling is essential for error correction and a further improvement in rendering quality.

\subsection{Discussion} In this section, we delve deeper into the aspects of DOT's rendering speed and training time.

\noindent \textbf{Doubled rendering speed. } As illustrated in \cref{fig:data_fig}, POT comprises a substantial number of voxel regions distributed in empty spaces and unnecessary divisions, leading to increased computation and memory consumption. In contrast, DOT exhibits more compact spatial divisions than POT, allowing the rays to bypass sparse and redundant areas.

\noindent \textbf{Negligible training time. } DOT solely calibrates  octrees extracted from the implicit NeRFs, resulting in a fast training process. The signal-based search algorithm scales linearly with the octree size. Therefore, as demonstrated in \cref{fig:abls}, it initially reduces more than $50\%$ of parameters, leading to enhanced efficiency due to the smaller size.

\subsection{Cross-device Analysis}
DOT is adaptable to scenarios with varied computational resources. 
We evaluate our method on the devices under distinct conditions, including A100-PCIE-40 GB, GeForce-RTX-3090-24GB, and GeForce-MX150-2GB. 
Specifically, MX150 was announced in mid-2017, tailored for mobile devices such as thin and light laptops. 
As is shown in \cref{tb:cs_devices}, we compare the run-time memory cost and the inference time for three devices with both full and compressed models.
Our models successfully accelerate the original models about 1.5 times FPS in the NeRF-Synthetic. 
For Tanks $\&$ Temples, it raises the rendering speed about 2-3 times. 
Notably, our full models perform at 11 FPS on MX150(only 2GB memory), while PlenOctree fails to launch. 
After compression, the memory reduction remains effective, bringing our models' access to more challenging datasets like the Tanks $\&$ Temples.
It proves DOT's ability to be applied more wildly in  web rendering, especially for situations with limited resources. 
\textit{We welcome all readers to refer to the recorded videos, including the real device tests and the visual quality comparisons in our suppl. materials.}

\section{Conclusion and Future Work}
We propose DOT, a dynamic structure to address the limitation of the fixed octree design in POT and allow for adaptive adjustment of the octree division for a more memory-efficient representation with higher quality. 
Compared with the original POT model, our method successfully shrinks over half of the model size, raises the rendering speed about one time, and even enhances the quality. 
DOT is buttressed by the hierarchical feature fusion strategy during the iterative rendering process, which maintains the globally consistent features instead of dropping them out. 
Moreover, our model can be applied to more scenarios with limited computational resources with flexible control over the strength of pruning/sampling operations.
However, our proposed method cannot reduce the excessive training time for its precursor NeRF-SH, which is required for both POT and DOT, from which they resample and cache the learned properties into the octree leaves for fast inference and optimization. 
In the future, we plan to explore extensions of our method, enabling the model training from scratch by methods such as reinforcement learning to automatically allocate samples with signal guidance to construct the 3D objects.

%
%
\noindent \textbf{Acknowledgement}: This paper is supported by the National Natural Science Foundation of China (NSF) under Grant No. NSFC22FYT45 and the Guangzhou City, University and Enterprise Joint Fund under Grant No. SL2022A03J01278.

\clearpage

{\small
\bibliographystyle{ieee_fullname}
\bibliography{egbib}
}

\clearpage
\appendix

\noindent{\large\textbf{Appendix}}

\section{Overview}
The algorithm of the DOT pipeline is illustrated in \cref{alg:the_alg}.
We provide further qualitative comparisons with the baseline PlenOctree\cite{yu2021plenoctrees} and our models with recursive pruning in \cref{fig:qual_cp}. 
We display more qualitative results of our methods on \cref{fig:full_syn} and \cref{fig:full_tt}. 
We also report the per-scene evaluation against PlenOctree in \cref{tab:s_sync} and \cref{tab:s_tt}.
Additionally, to examine the assumption that the neighboring features can be propagated, we carry out the control experiment in \cref{sec: assump}. Then we verify DOT's generalization ability with the case study described in \cref{sec:gel}.
Finally, more experiment details along the video demos are presented in \cref{sec:exp}.

\section{Additional Results}

\subsection{Qualitative Comparison}
\begin{figure*}[!t]
    \centering
    \includegraphics[width=0.8\linewidth]{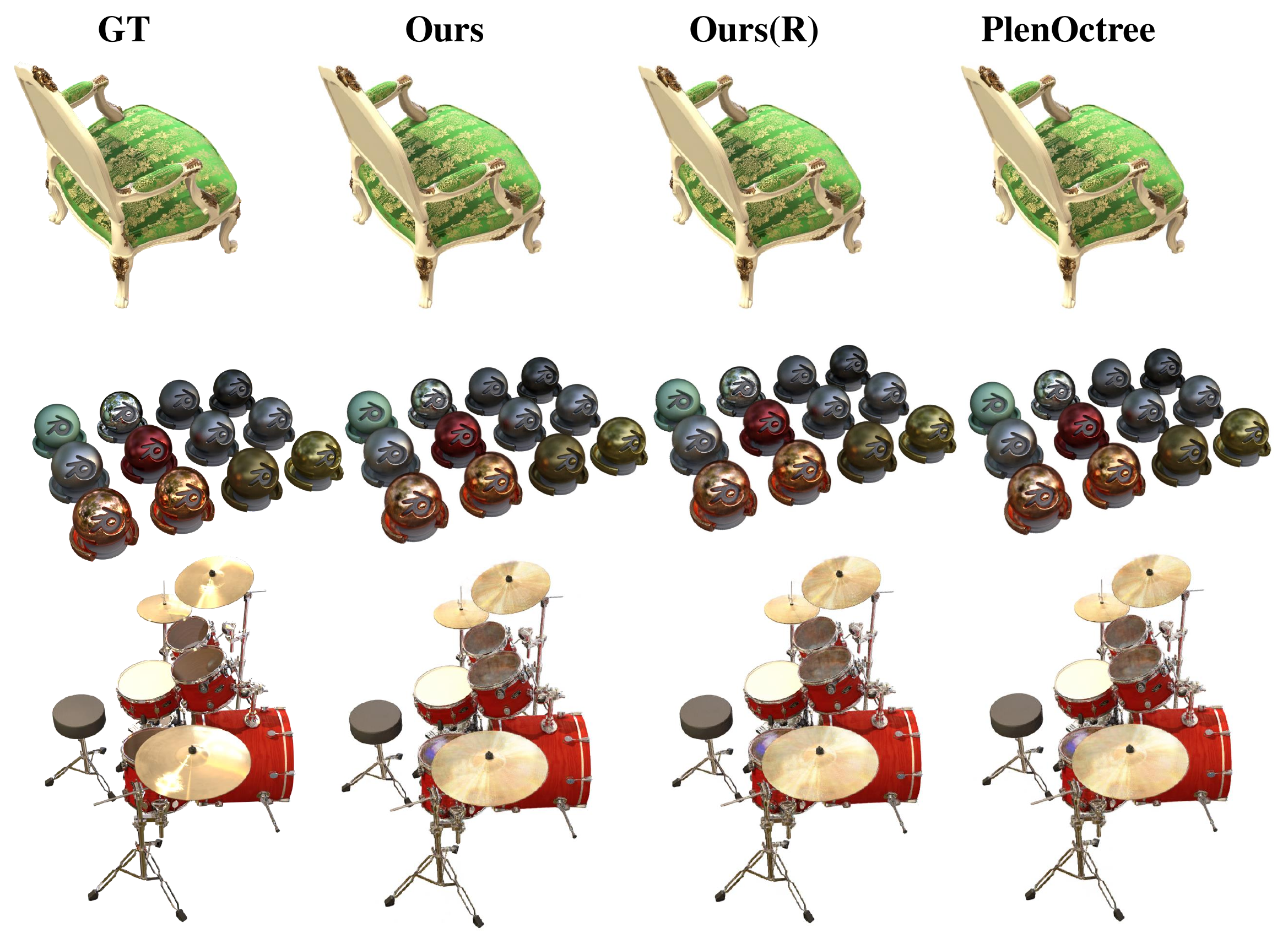}
    \caption{\textbf{Comparison on a list of models on NeRF-synthetic. } 
    }
    \label{fig:qual_cp}
\end{figure*}

In \cref{fig:qual_cp}, we compare our methods, including DOT denoted \emph{Ours}, DOT with recursive pruning \emph{Ours(R)}, \emph{PlenOctree} with the ground truth \emph{GT}. 
As observed in the rendering results, our methods show a similar visual quality to PlenOctree, while shrinking over half the memory sizes. Noticeably, compared with the ground truth images, both DOT and PlenOctree can provide satisfactory results regarding scenes such as \emph{chair} in the first row. However, for more challenging scenes \eg \emph{materials}, and \emph{drums} in the rest of the two rows, they cannot render the high-frequency regions \eg the reflection on the cymbal of the drums pleasingly.
The limitation may be addressed by increasing the sampling density on those regions using sampling operation like we have discussed in the main paper, splitting the scenes into the transmitted and reflected components\cite{Guo_2022_CVPR}, or using other physical-based rendering approximation\cite{ref-nerf}.

\begin{algorithm}[t]
\caption{DOT Model Training}
\label{alg:the_alg}
\begin{algorithmic}[1]
\State \textbf{Declaration}:
\State Rays $rays$, ground truth $GT$ and prediction $pred$
\State $DOT$ functions: $render(), merge()$ and $sample()$
\State Use recursive pruning $rec$, and the epoch number $i$
\State 
\Procedure{training}{$\tau, \gamma, rays, GT, i$}  
    \State $\textbf{Q}\leftarrow$ \textbf{optimization($rays, GT$)}
    \If{$i \mod T = 0$} 
    \State $\textbf{pruning}(\textbf{Q}, \tau)$
    \State \textbf{sampling}($\textbf{Q}, \gamma$)  
    \EndIf
\EndProcedure
\State 
\Procedure{optimization}{$rays, GT$}
    \State $rays \leftarrow permutate(rays)$
    \State $\textbf{Q}, pred \leftarrow DOT.render(rays)$ 
    \State $loss \leftarrow MSE(pred, GT)$
    \State return $\mathbf{Q}$
\EndProcedure

\State 
\Procedure{pruning}{$\textbf{Q}, \tau, rec$}:
  \While{$True$}
  \State $sel \leftarrow \mathbf{Q} \leq \tau$
    \State $merge(sel)$
  \If{$not \ rec \ \textbf{or} \ sel \ is \ empty$}
  \State break;
  \EndIf
  \EndWhile

\EndProcedure
\State
\Procedure{sampling}{$(\textbf{Q}, \gamma$}:
\State $sel \leftarrow topk(\gamma, Q)$
\State $sample(sel)$
\EndProcedure
\end{algorithmic}
\end{algorithm}

\subsection{Qualitative Evaluation}
\begin{figure*}[t]
    \centering
    \includegraphics[width=\linewidth]{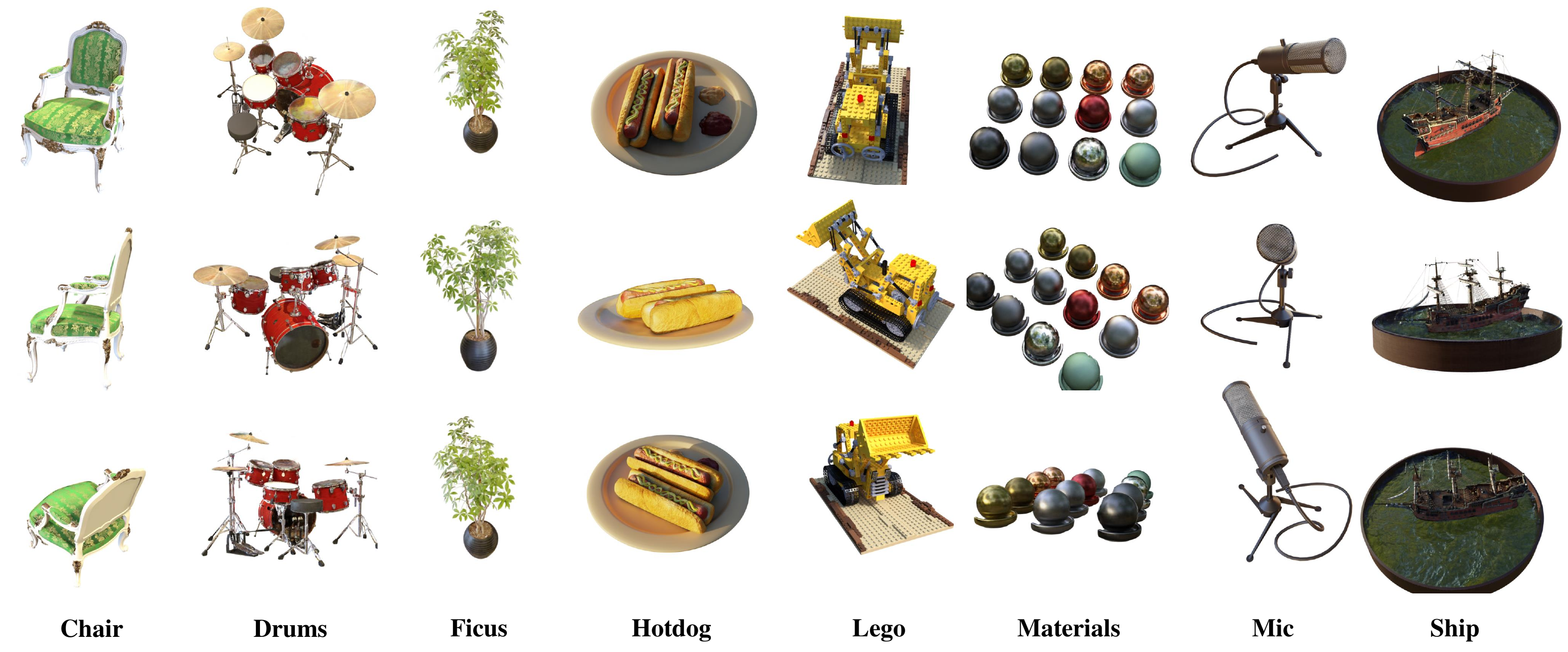}
    \caption{\textbf{More qualitative results of our DOT on NeRF-synthetic.} }
    \label{fig:full_syn}
\end{figure*}

\begin{figure*}[t]
    \centering
    \includegraphics[width=\linewidth]{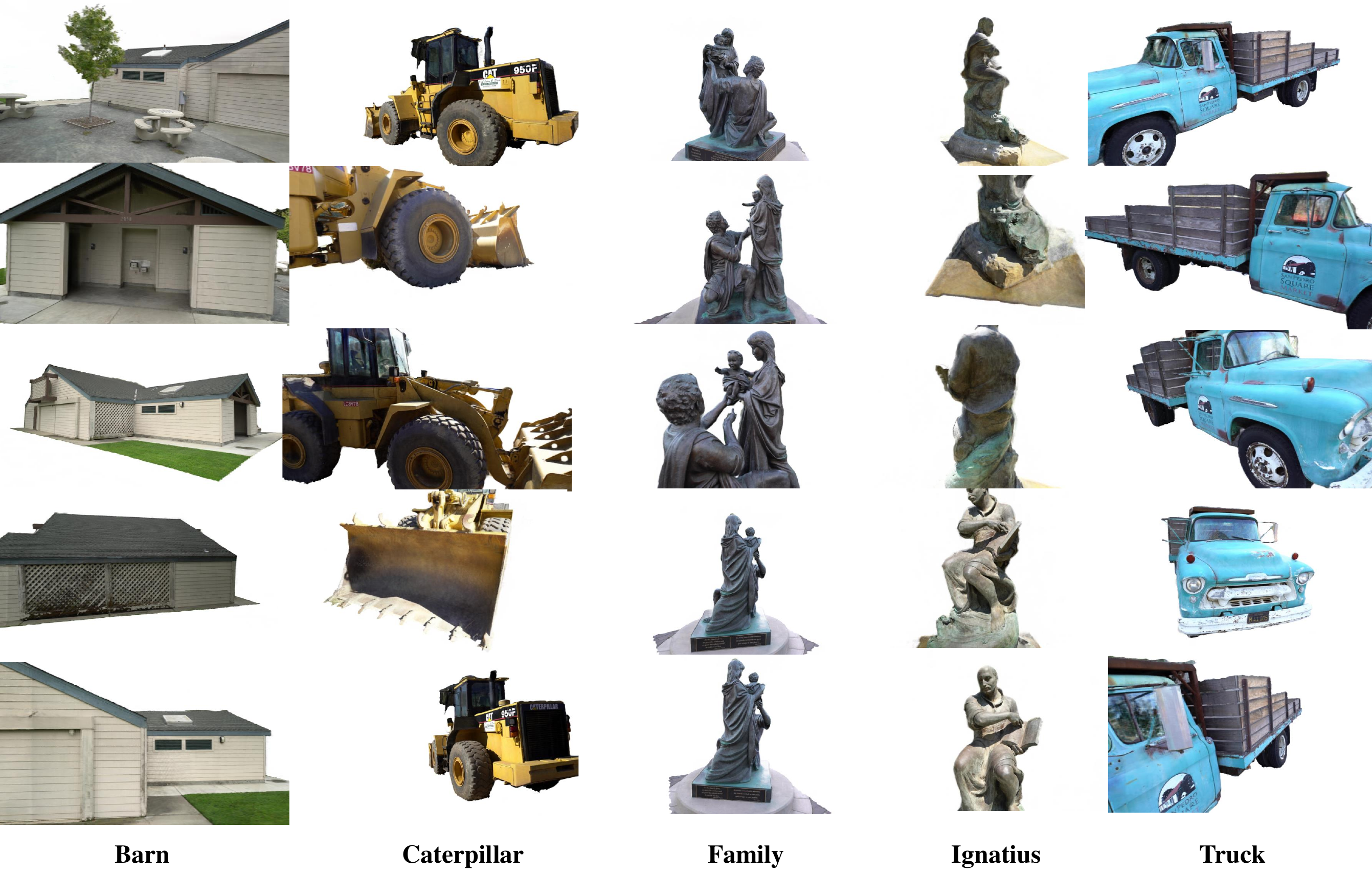}
    \caption{\textbf{More qualitative results of our DOT on Tanks\&Temples.} }
    \label{fig:full_tt}
\end{figure*}

We provide more qualitative results on both the NeRF-Synthetic(See \cref{fig:full_syn}) and the Tank \& Temples(See \cref{fig:full_tt}). 
We sample every scene in the datasets, using the poses picked randomly. 
All the samples are generated using the pre-trained DOT model \emph{Ours}. 
It turns out that the DOT is capable of rendering photo-realistic results.

\subsection{Per-scene Results}
We also provide the per-scene evaluation metrics on the NeRF-synthetic dataset (See \cref{tab:s_sync}) and Tank\&Template dataset (See \cref{tab:s_tt}). 

\begin{figure}[t!]
    \centering
    \includegraphics[width=0.75\linewidth]{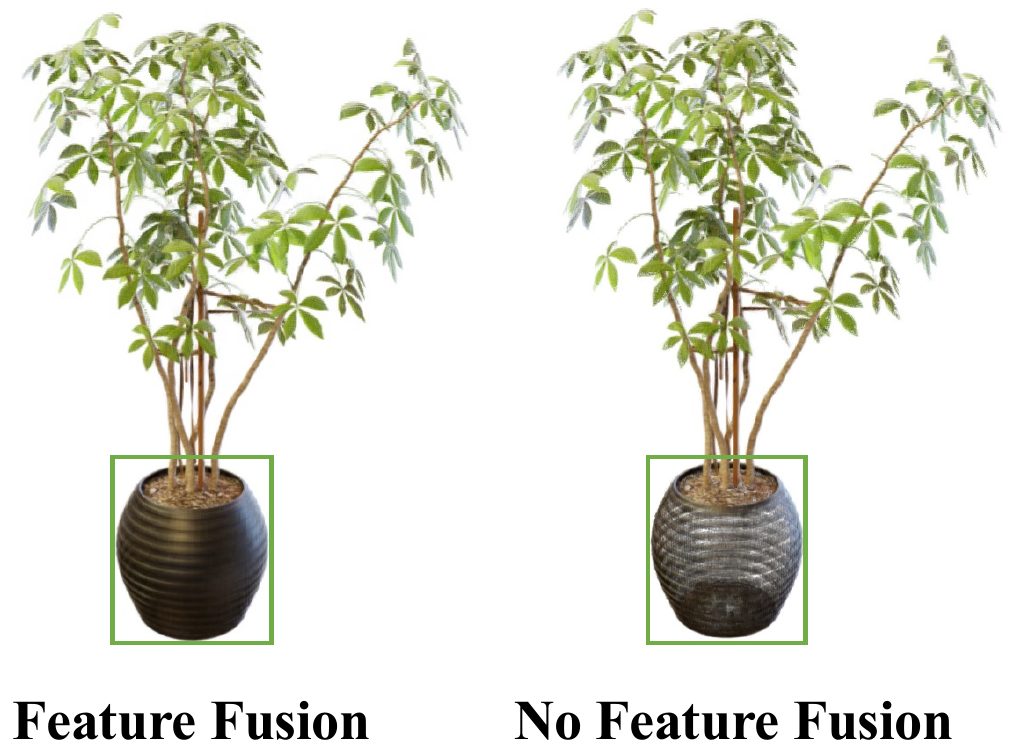}
    \caption{\textbf{The ablation study on the feature fusion.
    }
    We train the models on the \emph{ficus} scene of the synthetic dataset for 100 epochs by feature fusion and initializing properties to learn from scratch. 
    }
    \label{fig:abl_self_cp}
\end{figure}

\section{Technical Details}
\subsection{Assumption Verification}
\label{sec: assump}
To verify the neighboring assumption in Sec 3.2 of the main paper, we conduct a complementary experiment by disabling fusion (see \cref{fig:abl_self_cp}) in the training process. The loss of globally consistent features in the highlighted box confirms that the features are shared across octrees.
\subsection{Generalization Ability Verification}
\label{sec:gel}

\begin{figure*}[t]
    \centering
    \includegraphics[width=0.7\linewidth]{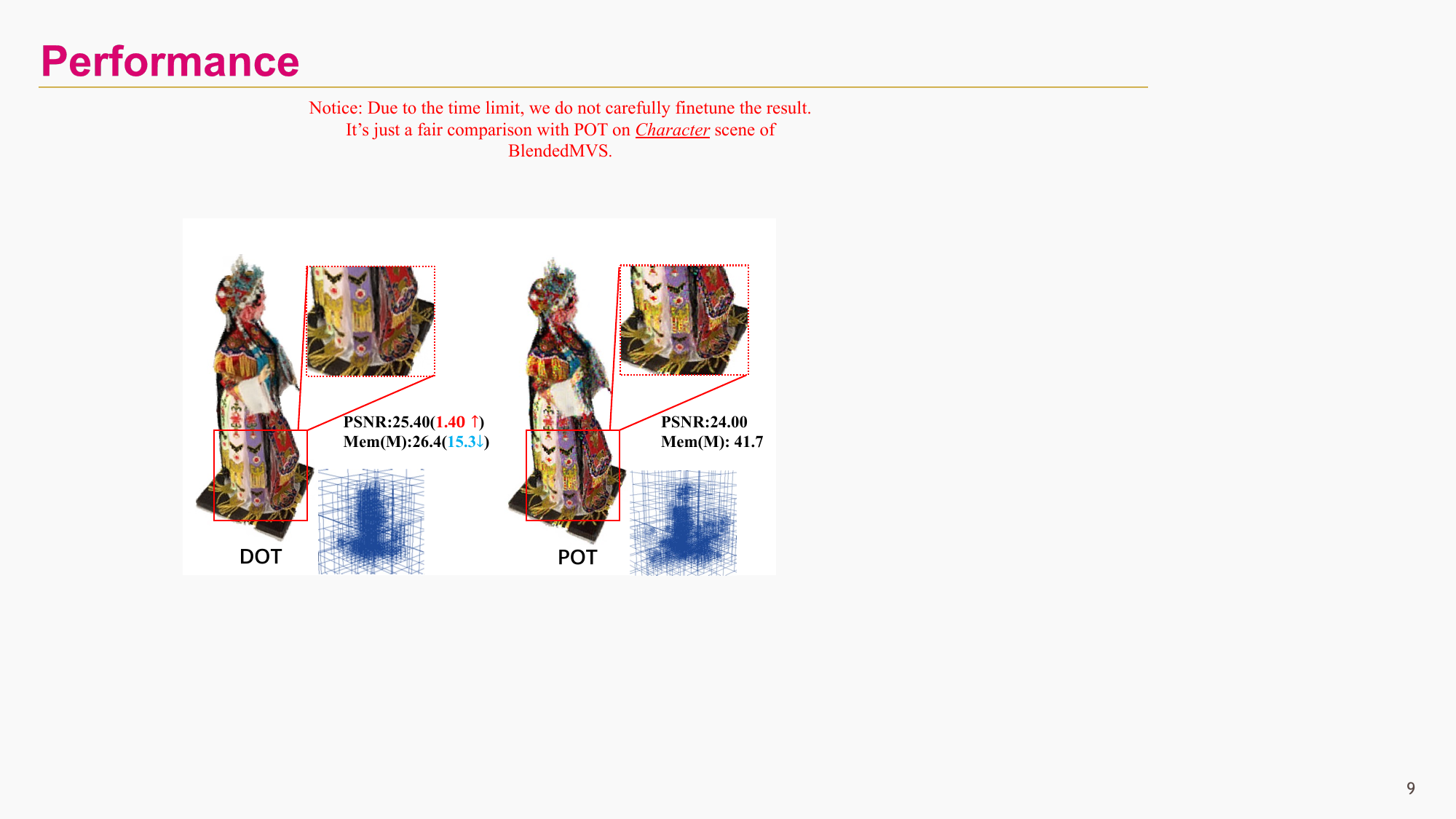}
    \caption{\textbf{The comparison between DOT and POT on \underline{Character} scene of BlendedMVS.} }
    \label{fig:character}
\end{figure*}

Although we target the \textbf{same} scenes as POT, the generalization ability for other NeRF datasets such as BlendedMVS\cite{liu2020neural} is also tested. 
As shown in \cref{fig:character}, the DOT outperforms POT with $+1.4$ PSNR, $-15.3$ MB memory, and a more compact octree structure. The zoomed boxes reveal that DOT provides much more intrinsic details. As we have included in our conclusion in the main paper, DOT can't reduce the excessive training time for its precursor NeRF-SH, which is required for both POT and DOT, from which they resample and cache the learned properties. We welcome interested readers to further the study on the generalization ability for more detests such as 360 data or the LLFF dataset. 

\subsection{Experiment Details}
\label{sec:exp}

\subsection{Device Information}
MX150 embedded in the laptop has 384 CUDA cores with a clock rate of 1468 MHz and a memory data rate of 6.01 Gbps. Its bandwidth is 48.06 GB per second, and its shared system memory is 4038 MB.  

\section{Videos Details}
The video attached to our supplementary materials consists of the following sections:

1) The general introduction about the octree representation.

2) The DOT's sampling and pruning operation demos. 

3) The overview of DOT's pipeline.

4) The Real machine testing on the laptop with MX150.

5) The visual comparison with PlenOctree.

6) The visual comparison between the model \emph{our}, \emph{our(R)} and their compressed models.

\clearpage

\begin{table*}[]
\centering
\caption*{PSNR$\uparrow$}
\vspace{-10pt}
\begin{tabular}{l|llllllll|l}
\hline
Methods & chair  & drums  & ficus  & hotdog & lego   & materials & mic    & ship & Avg \\  \hline
PlenOctree & 34.66   & 25.31 & 30.79 & 36.79  & 32.95  & 29.76     & 33.97 & 29.42 & 31.71  \\ \hline
ours(R) & 34.74   & 26.23 & 31.16  & 36.67  & 33.67  & 29.89     & 34.23 &  29.61 & 32.00  \\ \hline %
ours    & 34.82   & 26.25 & 31.16  & 36.76  & 33.82  & 30.24     & 34.24 & 29.61  & 32.11 \\ \hline 
\end{tabular}
\vspace{10pt}
\caption*{SSMI$\uparrow$}
\vspace{-10pt}
\begin{tabular}{l|llllllll|l}
\hline
Methods & chair  & drums  & ficus  & hotdog & lego   & materials & mic    & ship   & Avg    \\ \hline
PlenOctree & 0.9809 & 0.9330 & 0.9705 & 0.9822 & 0.9714 & 0.9549    & 0.9872 & 0.8841 & 0.9580 \\ \hline
ours(R) & 0.9807 & 0.9327 & 0.9716 & 0.9818 & 0.9705 & 0.9498    & 0.9875 & 0.8867 & 0.9577 \\ \hline
ours    & 0.9810 & 0.9329 & 0.9718 & 0.9823 & 0.9711 & 0.9547    & 0.9876 & 0.8868 & 0.9585 \\ \hline
\end{tabular}
\vspace{10pt}
\caption*{LPIPS$\downarrow$}
\vspace{-10pt}
\begin{tabular}{l|llllllll|l}
\hline
LPIPS   & chair  & drums  & ficus  & hotdog & lego   & materials & mic    & ship   & Avg    \\ \hline
PlenOctree & 0.0223 & 0.0764 & 0.0378 & 0.0319 & 0.0337 & 0.0593    & 0.0168 & 0.1441 & 0.0528 \\ \hline
ours(R) & 0.0225 & 0.0762 & 0.0397 & 0.0338 & 0.0348 & 0.0683    & 0.0171 & 0.1374 & 0.0537 \\ \hline
ours    & 0.0221 & 0.0764 & 0.0393 & 0.0317 & 0.0343 & 0.0621    & 0.0169 & 0.1372 & 0.0525 \\ \hline
\end{tabular}
\vspace{10pt}
\caption*{FPS$\uparrow$}
\vspace{-10pt}
\begin{tabular}{l|llllllll|l}
\hline
Methods       & chair & drums & ficus & hotdog & lego  & materials & mic   & ship  & Avg   \\ \hline
A100 PlenOctree  & 593.5 & 327.9       & 188.9  & 256.3 & 358.1    & 147.1 & 577.3 & 160.4 & 326.2 \\ \hline
A100 ours(R)  & 755.6 & 519.9       & 494.8  & 348.8 & 585.0    & 291.9 & 925.2 & 223.9 & 518.1 \\ \hline
A100 ours   & 753.0 & 490.0       & 414.0  & 327.0 & 553.8    & 276.2 & 870.9 & 213.0 & 487.2 \\ \hline  
3090 PlenOctree  & 550.2 & 302.4 & 117.4 & 78.8   & 279.1 & 59.8      & 561.4 & 51.7  & 250.1 \\ \hline
3090 ours(R)  & 744.3 & 468.4 & 536.4 & 196.6  & 571.7 & 249.1     & 929.5 & 97.4  & 474.2 \\ \hline
3090 ours     & 722.8 & 462.0 & 488.9 & 189.7  & 531.6 & 249.5     & 879.6 & 92.5  & 452.1 \\ \hline
MX150 PlenOctree & 15.38     & 7.59     & \XSolidBrush     & \XSolidBrush      & \XSolidBrush     & \XSolidBrush         & 12.81     & \XSolidBrush    & \XSolidBrush     \\ \hline
MX150 ours(R) & 19.72 & 11.56 & 11.19 & 7.15   & 13.56 & 5.95      & 19.74 & 3.90  & 11.60 \\ \hline
MX150 ours    & 18.87 & 10.67 & 9.92  & 6.70   & 12.43 & 5.75      & 18.19 & 3.70  & 10.78 \\ \hline
A100 PlenOctree*  & 675.1 & 344.0 & 166.7 & 204.1  & 347.5 & 145.0     & 610.7 & 141.0 & 329.3 \\ \hline

A100 ours(R)* & 927.5 & 587.9 & 486.5 & 373.0  & 690.2 & 318.8     & 908.0 & 231.1 & 565.4 \\ \hline 

A100 ours*    & 894.4 & 550.0 & 442.7 & 351.6  & 638.8 & 302.4     & 854.0 & 215.2 & 531.1 \\ \hline

3090 PlenOctree*  & 606.9 & 340.6       & 167.4  & 146.7 & 356.7    & 72.5  & 615.8 & 71.9  & 288.8 \\ \hline
3090 ours(R)*  & 765.4 & 501.3       & 545.1  & 211.9 & 593.0    & 252.7 & 967.6 & 101.5 & 492.3 \\ \hline
3090 ours*     & 733.8 & 477.2       & 502.5  & 207.5 & 557.1    & 252.4 & 909.0 & 97.7  & 467.2 \\ \hline
MX150 PlenOctree* & 15.90     & 7.85     & \XSolidBrush     & \XSolidBrush      & \XSolidBrush     & \XSolidBrush         & 13.62     & \XSolidBrush     & \XSolidBrush     \\ \hline
MX150 ours(R)* & 18.8  & 11.5        & 11.2   & 7.3   & 13.7     & 5.8   & 19.6  & 3.9   & 11.5  \\ \hline
Mx150 ours*    & 19.6  & 11.0        & 10.5   & 7.3   & 13.1     & 5.7   & 18.8  & 3.9   & 11.2  \\ \hline

\end{tabular}


\vspace{10pt}
\caption*{Checkpoint/Memory (GB)$\downarrow$}
\vspace{-10pt}
\begin{tabular}{c|cccccccc|c}
\hline
Methods & chair & drums & ficus & hotdog & lego & materials & mic  & ship & Avg  \\ \hline
PlenOctree & 0.81  & 1.3   & 1.8   & 2.7    & 2.1  & 3.7       & 0.43 & 2.7  & 1.94 \\ \hline
ours(R) & 0.56  & 0.60  & 0.41  & 1.29    & 0.81 & 0.91      & 0.26 & 1.53  & 0.80 \\ \hline
ours    & 0.58  & 0.65  & 0.47  & 1.38    & 0.91 & 1.01      & 0.28 & 1.70  & 0.87 \\ \hline
PlenOctree* & 0.19   & 0.28 & 0.43   & 0.43   & 0.10     & 0.26   & 0.88 & 0.56  & 0.39  \\ \hline
ours(R)* & 0.15   & 0.15 & 0.11   & 0.30   & 0.21     & 0.23   & 0.07 & 0.38  & 0.20  \\ \hline
ours* & 0.16   & 0.16 & 0.12   & 0.31   & 0.23     & 0.26   & 0.08 & 0.41 & 0.22 
 \\ \hline
\end{tabular}

\caption{Per-scene quantitive results on NeRF-synthetic dataset.* denotes the model is compressed}
\label{tab:s_sync}
\end{table*}

\begin{table*}[]
\centering
\caption*{PSNR$\uparrow$}
\vspace{-10pt}
\begin{tabular}{l|lllll|l}
\hline
Methods & Truck & Barn  & Caterpillar & Family & Ignatius & Avg   \\ \hline
PlenOctree & 26.84 & 26.80 & 25.29       & 32.85  & 28.20    & 28.00 \\ \hline
ours(R) & 27.05 & 27.45 & 25.58       & 32.94  & 28.22    & 28.25 \\ \hline
ours    & 27.08 & 27.43 & 25.59       & 33.05  & 28.25    & 28.28 \\ \hline
\end{tabular}

\vspace{10pt}
\caption*{SSMI$\uparrow$}
\vspace{-10pt}
\begin{tabular}{l|lllll|l}
\hline
Methods & Truck  & Barn   & Caterpillar & Family & Ignatius & Avg    \\ \hline
PlenOctree & 0.8559 & 0.9067 & 0.9622      & 0.9139 & 0.9480   & 0.9173 \\ \hline
ours(R) & 0.8689 & 0.9099 & 0.9633      & 0.9180 & 0.9494   & 0.9219 \\ \hline
ours    & 0.8691 & 0.9099 & 0.9644      & 0.9184 & 0.9488   & 0.9221 \\ \hline
\end{tabular}

\vspace{10pt}
\caption*{LPIPS$\downarrow$}
\vspace{-10pt}
\begin{tabular}{l|lllll|l}
\hline
Methods & Truck  & Barn   & Caterpillar & Family & Ignatius & Avg    \\ \hline
PlenOctree & 0.2259 & 0.1477 & 0.0691      & 0.1296 & 0.0802   & 0.1305 \\ \hline
ours(R) & 0.2067 & 0.1416 & 0.0640      & 0.1203 & 0.0750   & 0.1215 \\ \hline
ours    & 0.2070 & 0.1415 & 0.0612      & 0.1194 & 0.0761   & 0.1210 \\ \hline
\end{tabular}

\vspace{10pt}
\caption*{FPS$\uparrow$}
\vspace{-10pt}
\begin{tabular}{l|lllll|l}
\hline
Methods        & Barn   & Caterpillar & Family & Truck  & Ignatius & Avg    \\ \hline
A100 PlenOctree  & 94.44  & 113.47      & 77.12  & 129.90 & 40.41    & 91.07  \\ \hline 
A100 ours(R)     & 180.9 & 176.50       & 258.70  & 234.20 & 159.50    & 202.00 \\ \hline

A100 ours     & 159.93 & 167.09      & 225.55 & 222.33 & 108.59   & 176.70 \\ \hline
3090 PlenOctree   & 85.55  & 99.79       & 39.64  & 17.39  & 127.42   & 73.96  \\ \hline
3090 ours(R)   & 180.45 & 187.08      & 302.13 & 162.79 & 248.00   & 216.09\\ \hline
3090 ours       & 167.08 & 175.87      & 248.39 & 113.00 & 226.66   & 186.20 \\ \hline

MX150 PlenOctree & \XSolidBrush     & \XSolidBrush     & \XSolidBrush & \XSolidBrush      & \XSolidBrush     & \XSolidBrush  \\ \hline
MX150 ours(R) & \XSolidBrush     & \XSolidBrush     & \XSolidBrush & \XSolidBrush      & \XSolidBrush     & \XSolidBrush  \\ \hline
MX150 ours & \XSolidBrush     & \XSolidBrush     & \XSolidBrush & \XSolidBrush      & \XSolidBrush     & \XSolidBrush  \\ \hline

A100 PlenOctree*  & 131.4 & 129.5       & 87.70   & 145.50 & 68.40     & 112.50 \\ \hline

A100 ours(R)* & 177.97 & 178.90      & 281.52 & 245.00 & 156.20   & 207.92 \\ \hline
A100 ours*   & 170.40 & 178.33 & 235.26 & 234.23 & 116.00 & 183.00\\ \hline

3090 PlenOctree*  & 85.5  & 99.8        & 39.6   & 17.4  & 127.4    & 73.9  \\ \hline
3090 ours(R)* & 348.0 & 379.0       & 150.0  & 275.0 & 196.0    & 269.6 \\ \hline
3090 ours*      & 322.0 & 349.0       & 117.0  & 239.0 & 166.0    & 238.6 \\ \hline

MX150 PlenOctree* & \XSolidBrush     & \XSolidBrush     & \XSolidBrush & \XSolidBrush      & \XSolidBrush     & \XSolidBrush  \\ \hline
MX150 ours(R)* & 2.92   & 2.85        & 4.60   & 2.55   & 3.89     & 3.36   \\ \hline
MX150 ours*    & 2.85   & 2.68        & 3.87        & 1.90       & 3.61        & 2.98      \\ \hline
\end{tabular}


\vspace{10pt}
\caption*{Checkpoint/Memory (GB)$\downarrow$}
\vspace{-10pt}
\begin{tabular}{l|lllll|l}
\hline
Methods & Truck & Barn & Caterpillar & Family & Ignatius & Avg   \\ \hline
PlenOctree & 2.5   & 2.4  & 2.9         & 2.3    & 3        & 2.62  \\ \hline
ours(R) & 1.3   & 1.4  & 0.4         & 0.87   & 0.64     & 0.92 \\ \hline
ours    & 1.5   & 1.5  & 0.53        & 1.1    & 0.85     & 1.10 \\ \hline
PlenOctree* & 0.60   & 0.67        & 0.84   & 0.53   & 0.63     & 0.65  \\ \hline
ours(R)* & 0.25   & 0.34        & 0.37   & 0.12   & 0.17     & 0.25   \\ \hline
ours* & 0.29   & 0.37        & 0.40   & 0.16   & 0.21     & 0.29   \\ \hline
\end{tabular}

\caption{Per-scene quantitive results on Tanks\&Temples dataset.* denotes the model is compressed}
\label{tab:s_tt}
\end{table*}

\end{document}